\documentclass[a4paper, 10pt, conference]{ieeeconf}      

\IEEEoverridecommandlockouts                              

\usepackage{graphicx} 
\usepackage{amssymb}  
\usepackage{hyperref}
\usepackage{balance}
\usepackage{pifont} 
\newcommand{\xmark}{\ding{55}}

\title{\LARGE \bf
Peg-in-Bench: A Modular Benchmark for High-Precision Robotic Insertion
}

\author{Yosel Delgado*\textsuperscript{\textdagger}$^{1}$, Jos\'e G. Buenaventura-Carre\'on\textsuperscript{\textdagger}$^{1}$, Floris Erich\textsuperscript{\textdagger}$^{1}$, Roman Mykhailyshyn$^{1}$, \\Tomohiro Motoda$^{1}$, Koshi Makihara$^{1}$ and Yukiyasu Domae$^{1}$
\thanks{$^{*}$Corresponding author, reachable at yosel.delgado@aist.go.jp.}%
\thanks{$^{\textsuperscript{\textdagger}}$These authors contributed equally to this work and share first authorship.}%
\thanks{$^{1}$Yosel Delgado, Jos\'e Gustavo Buenaventura-Carre\'on, Floris Erich, Roman Mykhailyshyn, Tomohiro Motoda, Koshi Makihara and Yukiyasu Domae are with the National Institute of Advanced Industrial Science and Technology (AIST), Japan.}%
}

\begin{document}

\maketitle
\pagestyle{empty}

\begin{abstract}

High-precision insertion remains a fundamental challenge in robotic manipulation due to the strict alignment requirements and contact-rich interactions involved. Although peg-in-hole tasks are widely used for evaluation, existing benchmarks often rely on fixed task configurations, limiting their ability to assess robustness and generalization across different insertion scenarios. This paper introduces a reconfigurable peg-in-hole benchmark designed to evaluate task generalization in high-precision insertion. The benchmark consists of a set of fully 3D-printable modular components, including multiple peg geometries, tolerance levels, and configurable base structures that can be combined to generate a large variety of insertion and assembly tasks. By varying object layouts, orientations, and task structures while maintaining controlled physical conditions, the benchmark enables systematic evaluation of adaptation to unseen scenarios. To support reproducibility, we additionally provide a scenario generation tool capable of producing standardized task configurations and machine-readable task descriptions. The scenario generation tool and the STL files of the benchmark pieces are available through the project repository: \url{https://github.com/aistairc/peg-in-bench}.

\end{abstract}

\section{INTRODUCTION}

Peg-in-hole insertion is a fundamental problem in robotic assembly and contact-rich manipulation, serving as a representative task for evaluating perception, motion planning, and precision control under geometric constraints~\cite{kimble2020,mason2001}. To support reproducible research and meaningful comparison across robotic systems, benchmark protocols and standardized object sets have become increasingly important, with resources such as the YCB Object and Model Set playing a significant role in the manipulation community~\cite{calli2015ycb}. 

Several benchmarks have been proposed to assess robotic assembly capabilities, ranging from early assembly-oriented benchmarks such as the Cranfield benchmark and the Delft Assembly Test (DATE)~\cite{collins1985,heemskerk1988} to more recent frameworks for evaluating robotic learning and generalization~\cite{james2020rlbench,liu2023libero}. While these efforts have improved standardization and reproducibility, most peg-in-hole evaluations continue to rely on fixed task configurations, limiting their ability to assess how insertion skills transfer across different task layouts, object orientations, and assembly structures.

This limitation is particularly relevant as modern manipulation systems increasingly rely on learning-based approaches capable of adapting to previously unseen situations~\cite{james2020rlbench,liu2023libero}. Evaluating such capabilities requires benchmarks that go beyond measuring insertion success in a single configuration and instead allow systematic variation of task conditions while maintaining controlled and reproducible experimental settings. However, there remains a lack of physical benchmarks specifically designed to study task-level generalization in high-precision insertion.
To address this gap, we introduce \textit{Peg-in-Bench}, a modular and fully 3D-printable benchmark for high-precision peg-in-hole insertion. Unlike existing benchmarks that primarily evaluate performance on predefined task instances, Peg-in-Bench enables controlled variation of peg geometry, insertion tolerance, target position, target orientation, and assembly structure through a common set of reconfigurable components. This design allows a large number of reproducible insertion scenarios to be generated from the same physical components, supporting the evaluation of spatial, geometric, and task-compositional generalization. The benchmark is designed to remain accessible and inexpensive to deploy while supporting manipulation approaches ranging from classical compliant control to modern learning-based methods~\cite{luo2019variable,luo2021forceguided}.

\section{RELATED WORK}

\subsection{Benchmarking Robotic Manipulation and Assembly} 

Benchmarking has played a central role in advancing robotic manipulation by providing standardized tasks, evaluation procedures, and physical artifacts that enable reproducible comparison across research groups. Early efforts such as the YCB Object and Model Set established common object collections for manipulation research \cite{calli2015ycb}, while RLBench introduced a large-scale framework for evaluating a variety of robotic skills in simulation \cite{james2020rlbench}. More recently, cloud-based initiatives such as OCRTOC \cite{liu2021ocrtoc} and large-scale infrastructures such as ManipulationNet \cite{manipulationnet2026} have emphasized standardization and reproducibility through shared hardware, protocols, and evaluation procedures. 

These benchmarks have significantly improved the comparability of robotic manipulation systems. However, they primarily focus on evaluating performance within predefined task sets, offering limited support for systematically studying how insertion skills transfer across different physical task configurations.

\subsection{Generalization-Oriented Manipulation Benchmarks} 

Recent benchmark development has increasingly focused on evaluating generalization rather than performance on a single task instance. The Functional Manipulation Benchmark (FMB) provides a collection of procedurally generated assembly tasks designed to study object and task generalization in robotic learning \cite{luo2025fmb}. Similarly, FurnitureBench investigates long-horizon manipulation through realistic furniture assembly tasks \cite{heo2023furniturebench}, while Open X-Embodiment promotes cross-platform evaluation through large-scale robot learning datasets and shared learning resources \cite{openx2023}. 

These efforts highlight the importance of evaluating transfer across objects, environments, and manipulation procedures. A similar trend can be observed in recent vision-language-action systems such as RT-2, which demonstrate knowledge transfer across robotic tasks using large-scale pretraining \cite{zitkovich2023rt2}. However, they primarily target learning-based systems and do not provide a dedicated framework for studying the effects of insertion geometry, tolerance variation, and task reconfiguration on high-precision peg-in-hole manipulation.

\subsection{Robotic Insertion Benchmarks and Evaluation} 
Peg-in-hole insertion has traditionally served as a representative benchmark for contact-rich robotic manipulation and robotic assembly. Historical benchmarks such as the Cranfield benchmark \cite{collins1985} and the Delft Assembly Test (DATE) \cite{heemskerk1988} provided standardized assembly tasks to evaluate robotic precision and dexterity. More recently, the NIST Assembly Task Board introduced a broader collection of insertion and fastening operations representative of industrial assembly processes \cite{kimble2020}. QBIT further extended insertion benchmarking by incorporating quality-oriented metrics such as force energy and force smoothness, enabling more comprehensive assessment beyond task success \cite{schempp2025qbit}. 

Despite these contributions, existing insertion benchmarks generally rely on fixed task configurations. As a result, they provide limited support for evaluating whether insertion skills learned in one configuration transfer to unseen spatial layouts, orientations, or assembly structures.

\subsection{Learning-Based Insertion Frameworks}

Parallel to benchmark development, several learning-based approaches have been proposed to improve insertion performance and generalization. InsertionNet introduced a scalable framework that combines visual and force information for peg-in-hole insertion \cite{spector2021insertionnet}, while EasyInsert demonstrated strong zero-shot transfer to unseen insertion objects through relative pose prediction and efficient real-world data collection \cite{li2025easyinsert}. Additional research has explored learning-based force-guided assembly and high-precision robotic insertion using reinforcement learning and imitation learning approaches \cite{luo2019variable,luo2021forceguided,wang2021assembly}. 

These methods have shown that robotic insertion can generalize across object geometries and environmental conditions when appropriate learning strategies are employed. However, evaluation is often performed using custom experimental setups and limited collections of insertion tasks. Consequently, comparing insertion policies across different studies remains difficult, reinforcing the need for standardized and configurable physical benchmarks.

\begin{table*}[t] 
\centering 
\caption{Comparison of representative insertion and manipulation benchmarks.}
\label{tab:benchmark_comparison} 
\begin{tabular}{lcccccc} 
\hline 
Benchmark & 
Precision Insertion & 
Tolerance Variation & 
Task Reconfiguration & 
Assembly Tasks & 
Primary Evaluation Focus \\ 
\hline 
NIST Assembly Task\cite{kimble2020} &
\checkmark & 
\xmark & 
\xmark & 
\checkmark & 
Assembly Operations \\ 
ManipulationNet \cite{manipulationnet2026} &
\checkmark & 
\checkmark & 
\xmark & 
\checkmark & 
Standardized Evaluation \\ 
FMB \cite{luo2025fmb} & 
\xmark & 
\xmark & 
\xmark & 
\checkmark & 
Object Generalization \\ 
EasyInsert \cite{li2025easyinsert} & 
\checkmark & 
\xmark & 
\xmark & 
\xmark & 
Policy Generalization \\ 
QBIT \cite{schempp2025qbit} & 
\checkmark & 
\xmark & 
\xmark & 
\xmark & 
Quality-Aware Evaluation \\ 
REASSEMBLE \cite{sliwowski2025reassemble} &
\checkmark & 
\xmark & 
\xmark & 
\checkmark & 
Assembly Learning Dataset \\ 
\textbf{Peg-in-Bench (Ours)} & 
\checkmark & 
\checkmark & 
\checkmark & 
\checkmark & 
Task Generalization \\ 
\hline 
\end{tabular} 
\end{table*}

\subsection{Comparison with Existing Benchmarks} 

Table~\ref{tab:benchmark_comparison} summarizes the characteristics of representative manipulation and insertion benchmarks. Existing benchmarks primarily focus on standardized performance evaluation \cite{liu2021ocrtoc,manipulationnet2026}, manipulation learning and object generalization \cite{luo2025fmb,heo2023furniturebench,openx2023}, or insertion-specific algorithm evaluation \cite{schempp2025qbit,spector2021insertionnet,li2025easyinsert}. In contrast, Peg-in-Bench is designed around task-level generalization in high-precision peg-in-hole insertion. Rather than defining a fixed collection of benchmark tasks, the proposed framework enables controlled variation of peg geometry, insertion tolerance, target position, target orientation, and assembly structure using a common set of reusable components. This design allows researchers to generate a large number of reproducible insertion scenarios while systematically evaluating a robot's ability to transfer insertion skills across previously unseen task configurations. As a result, the benchmark complements existing evaluation frameworks by focusing on how insertion capabilities generalize across task variations rather than on performance within a single predefined benchmark setting. 

The benchmark therefore targets three complementary forms of generalization: spatial generalization through changes in object position and orientation, geometric generalization through variation of peg shape and insertion tolerance, and task-compositional generalization through the construction of increasingly complex assembly structures. These dimensions constitute the primary design objectives of the proposed framework.

\section{BENCHMARK DESIGN}

The benchmark builds upon the peg geometries and tolerance specifications introduced in the insertion task of ManipulationNet~\cite{manipulationnet2026}, using their STL models as a foundation for our own component designs. Our primary objective was to create a benchmark that is both highly accessible and highly generalizable, enabling researchers to generate a large number of insertion scenarios using a small set of modular, reconfigurable pieces. This design philosophy allows the same physical components to be rearranged into different spatial configurations, orientations, and task layouts, supporting broad evaluation of robot precision, adaptability, and generalization. 

Our design philosophy is also inspired by previous benchmark efforts that employ reproducible 3D-printed components to study manipulation generalization and assembly performance~\cite{luo2025fmb},~\cite{heo2023furniturebench}. The modularity of the benchmark allows individual manipulation skills to be evaluated in isolation or composed into more complex multi-stage assembly scenarios, similar to approaches explored in long-horizon manipulation benchmarks~\cite{luo2025fmb},~\cite{heo2023furniturebench}.

\subsection{Design Objectives and Generalization Dimensions} 

The primary objective of Peg-in-Bench is to evaluate task-level generalization in high-precision peg-in-hole insertion. In practical manipulation systems, successful insertion requires more than achieving high performance on a single task instance. Instead, robots must adapt their insertion behavior when task conditions change while preserving the underlying insertion skill. For this reason, the benchmark was designed around three complementary forms of generalization. 

\textit{Spatial generalization} concerns to the ability of a robot to perform insertions at previously unseen target positions and orientations. In industrial applications, insertion targets are rarely encountered in exactly the same pose, and small spatial changes can significantly alter the alignment process. To evaluate this capability, the benchmark allows systematic variation of hole position and orientation through the modular base structure and the rotatable hole pieces. 

\textit{Geometric generalization} refers to the transfer of insertion strategies across different object geometries and clearance conditions. Different peg shapes introduce distinct alignment constraints and contact dynamics, while tolerance variations directly influence the precision required for successful insertion. The benchmark therefore incorporates multiple peg geometries and three tolerance levels to study the effect of geometric complexity and insertion precision on performance. 

\textit{Task-compositional generalization} focuses on the ability to combine insertion skills into larger manipulation objectives. While many benchmarks evaluate isolated peg-in-hole operations, real assembly tasks often require multiple insertions performed as part of a sequential procedure. The modular connection mechanism of the benchmark allows the construction of larger assembly structures that require the composition of several insertion actions within a single task. 

The benchmark focuses on these three dimensions because they directly affect the planning, perception, and contact interactions involved in insertion. Other sources of variation, such as illumination, camera placement, sensor noise, friction coefficients, material properties, or surface finish, were intentionally held approximately constant in the first version of the benchmark. Although these factors are relevant in practical applications, they primarily evaluate perception robustness or contact-model uncertainty. By controlling these variables, the benchmark isolates the influence of task configuration on insertion performance and enables more systematic comparison between manipulation approaches.

\subsection{Benchmark components}

The benchmark consists of three main component families: pegs, shaped‑hole pieces, and bases. To ensure dimensional accuracy, critical for evaluating high‑precision insertion, we manufacture the pegs and hole pieces using a resin printer with a 0.05 mm resolution. We recommend printing these two component types using different resin colors to improve visual distinguishability during perception‑based experiments. The bases and additional components, which do not require sub‑millimeter accuracy, can be produced using any standard filament‑based 3D printer.

\subsubsection{Pegs} The benchmark includes five peg geometries: circular, rectangular, hexagonal, triangular, and L‑shaped. These shapes were selected to span a range of insertion challenges, from symmetric geometries that simplify alignment to asymmetric or keyed shapes that require precise orientation control. This diversity enables evaluation of robot performance across tasks with varying geometric complexity and contact dynamics, supporting studies of shape‑dependent alignment and generalization across geometries. 

\subsubsection{Shaped-hole pieces} Following the tolerance specifications in ManipulationNet, we adopt three tolerance levels: 0.1 mm, 1 mm, and 3 mm. The smallest tolerance of 0.02 mm is excluded because it cannot be reliably reproduced using widely available 3D‑printing technologies. With five peg shapes and three tolerances, the benchmark includes 15 shaped‑hole pieces. Each hole piece is designed as an octagonal prism, providing eight distinct orientations. The hole is intentionally positioned off‑center, ensuring that each orientation produces a unique insertion configuration. To simplify the determination of the current orientation of each hole piece, a red line was added originating from one of its vertices. This visual marker provides an unambiguous orientation reference.

To ensure reliable positioning, each shaped-hole piece contains a N50 8 × 3 mm circular magnet embedded in its base, which mates with a corresponding magnet in the base pieces. The magnets provide sufficient retention force to keep the hole pieces fixed during normal insertion operations while enabling rapid replacement and reconfiguration. Unlike a rigidly fixed connection, the magnetic attachment can release when subjected to excessive lateral forces caused by failed insertions or incorrect extraction trajectories. As a result, the benchmark is less susceptible to breakage of the printed parts and unintended displacement of the full assembly, improving its robustness during repeated experimentation.

\subsubsection{Bases} The base pieces serve as modular holders for the shaped‑hole components. They allow users to easily replace, rotate, and reposition hole pieces to generate new tasks. Each base is a cube with interlocking joints and holes on its sides, enabling multiple bases to be connected into larger structures. A small indentation marks the canonical orientation of each base, and a magnet embedded in the bottom ensures secure attachment to the shaped‑hole pieces.

We designed four base variants, each supporting different structural configurations:

\begin{itemize}

\item Corner‑base — joints on the upper and right sides; holes on the lower and left sides.
\item Empty‑base — holes on all four sides.
\item Line‑base — holes on left and right sides; joints on upper and lower sides.
\item Cross‑base — joints on all four sides.

\end{itemize}

Beyond supporting peg‑in‑hole tasks, the bases also enable multi‑object insertion tasks. The joints and holes are manufactured with a 0.25 mm tolerance, providing a strong yet manipulable connection between bases. This allows robots to perform insertion using the bases themselves, expanding the benchmark beyond single‑peg tasks and enabling evaluation of simultaneous multi‑object manipulation.

\begin{figure}

\centering
{\label{fig:3dmodel}
\centering
\includegraphics[width=0.4\linewidth]{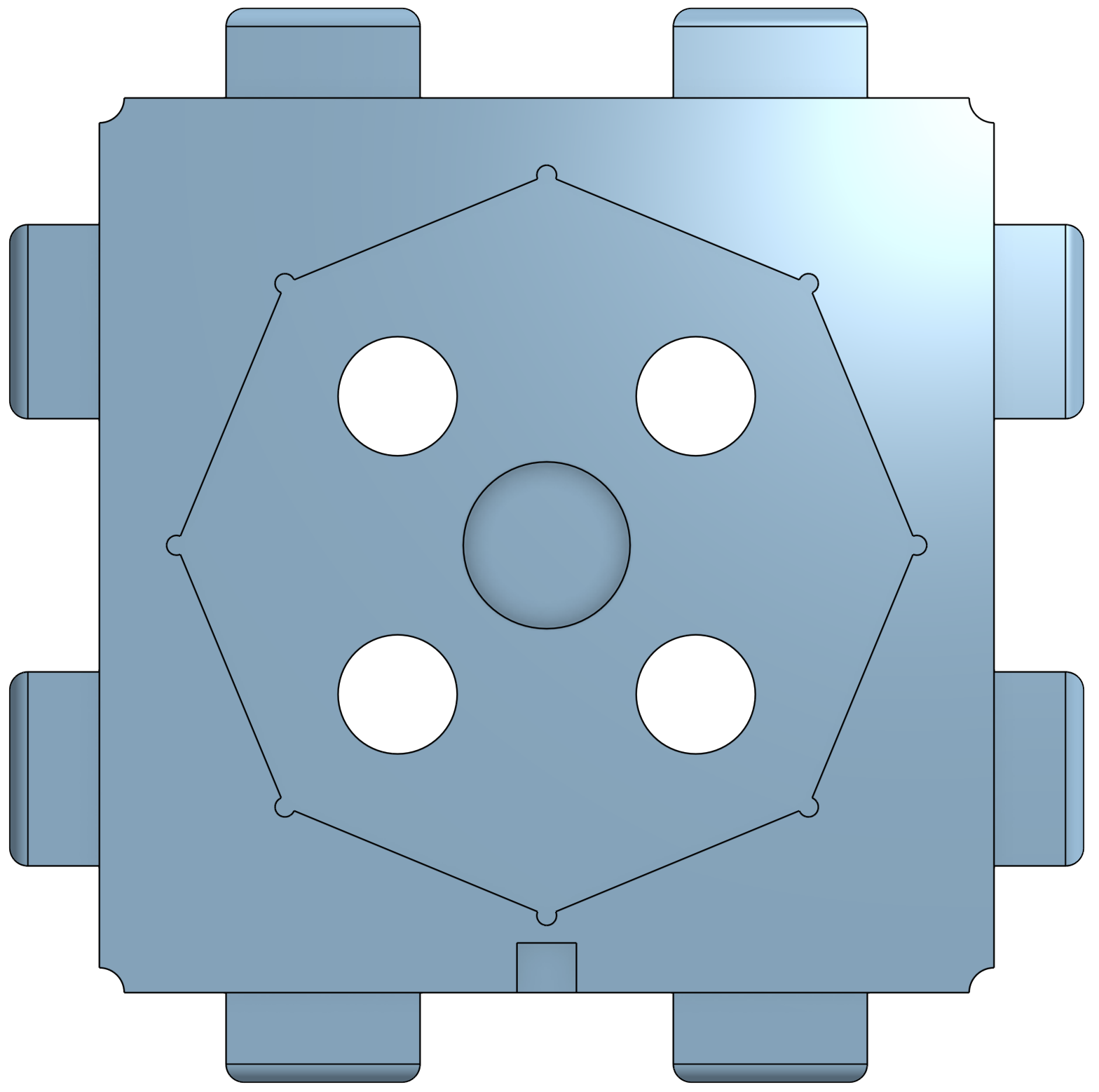}
}
\caption{Cross-base model on OnShape, showing the octagonal space to hold the shaped-hole pieces and the space to embed a magnet on the bottom.}
\end{figure}

\subsubsection{Additional pieces}

The benchmark includes two auxiliary components: a Peg‑holder and fixing pieces. The Peg‑holder provides a consistent initial pose for pegs across experiments. Its geometry mirrors that of the base pieces, allowing it to connect seamlessly using the same joint system. The fixing pieces are designed to secure bases to the workspace using M8 screws, ensuring stable positioning during task execution and preventing accidental displacement during robot interaction. 

\begin{figure}

\centering
{\label{fig:benchmarkpieces}
\centering
\includegraphics[width=0.85\linewidth]{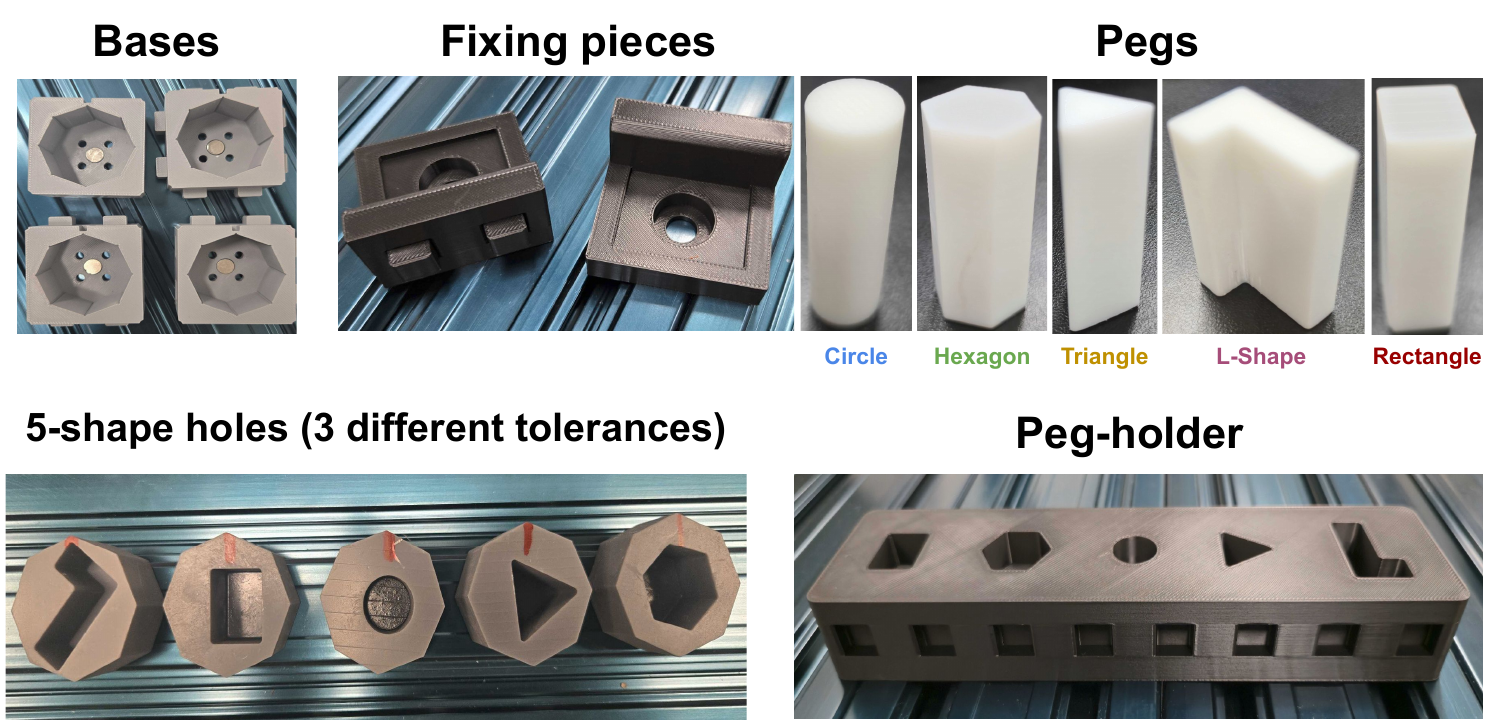}
}
\caption{Benchmark pieces printed using 3D printers.}
\end{figure}

\section{Benchmark tasks and evaluation scenarios}

Unlike traditional insertion benchmarks that rely on fixed task instances, Peg-in-Bench enables the generation of families of related tasks that systematically vary geometry, tolerance, spatial layout, and assembly structure. The following examples illustrate representative evaluation scenarios supported by the benchmark. These tasks can be structured to evaluate generalization across shapes, tolerances, workspace arrangements, and manipulation objectives. 

\begin{figure}

\centering
{\label{fig:Diagram}
\centering
\includegraphics[width=0.85\linewidth]{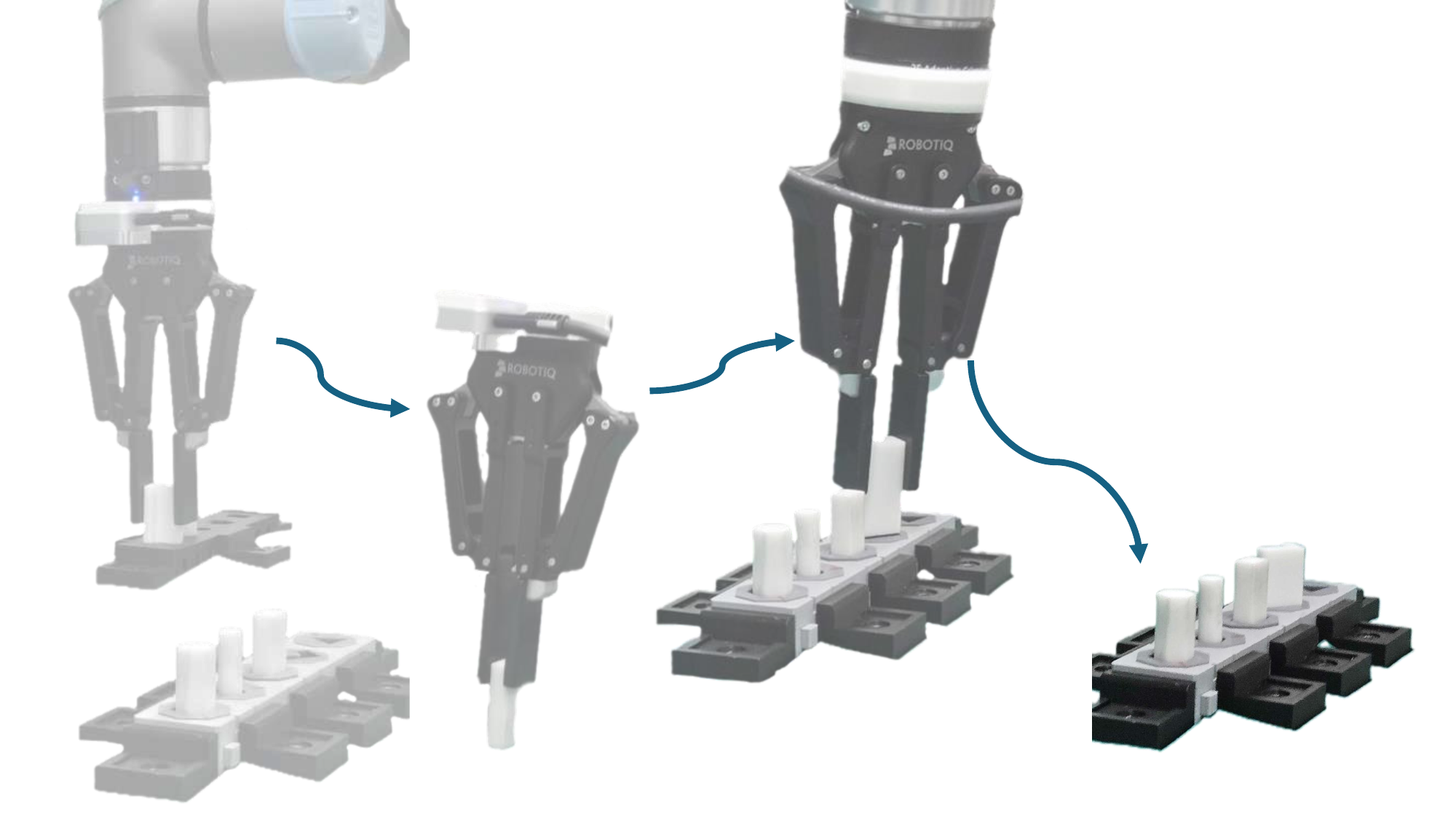}
}
\caption{Demonstration of the single-peg insertion task using the benchmark with a UR7e robot and Robotiq gripper.}
\end{figure}

\subsection{Single-peg insertion and shape generalization}

Single-peg insertion tasks provide a standardized way to evaluate precision manipulation and have been widely adopted in both classical assembly systems and modern learning-based insertion methods~\cite{kimble2020},~\cite{li2025easyinsert},~\cite{spector2021insertionnet}. The most basic use of the benchmark is evaluating insertion performance across different peg geometries. In this setting, the robot is required to insert all five peg shapes into their corresponding holes while maintaining a fixed tolerance level. The initial poses of the pegs remain constant, whereas the locations and orientations of the target holes are varied between scenarios.

Similar insertion tasks are found in benchmarks such as the NIST Assembly Task Board and ManipulationNet~\cite{kimble2020,manipulationnet2026}, however, in Peg-in-Bench the same task can be instantiated using multiple peg geometries and spatial configurations, allowing the evaluation of geometric and spatial generalization in addition to insertion performance.

\subsection{Tolerance-dependent insertion}

The benchmark can also be used to assess performance under varying insertion clearances. For a selected peg geometry, the robot is required to perform insertions using progressively smaller tolerances. Since the hole orientation can be randomized independently for each trial, the task simultaneously evaluates geometric alignment accuracy and the robot's capability to deal with increasingly restrictive contact conditions.

Controlled tolerance variation has been widely used to evaluate insertion precision in previous benchmarks. In Peg-in-Bench, tolerance changes can be combined with variations in target position and orientation, enabling the assessment of both insertion accuracy and generalization across task configurations~\cite{kimble2020},~\cite{luo2019variable},~\cite{luo2021forceguided}.

\subsection{Modular assembly tasks}

By combining multiple insertion operations into a larger assembly objective, the benchmark can also be used to evaluate skill composition, long-horizon manipulation, and task sequencing capabilities~\cite{luo2025fmb},~\cite{sliwowski2025reassemble},~\cite{heo2023furniturebench}, as well as assembly methods that combine force and trajectory learning for contact-rich manipulation~\cite{wang2021assembly}. The base pieces allow the benchmark to extend beyond conventional peg-in-hole problems into assembly operations. Robots can be tasked with connecting multiple base modules through the joint interfaces to construct larger structures. 

Simple assembly configurations involve connecting adjacent corner and empty bases to create linear arrangements, while more advanced configurations require the construction of irregular structures using line and cross bases. Because each connection itself is an insertion task, these experiments evaluate dexterous manipulation, spatial reasoning, and sequential planning.

Unlike assembly benchmarks with predefined procedures, the modular structure allows new assembly scenarios to be generated from the same set of components, supporting the evaluation of task-compositional generalization.

\section{Scenario generator tool}

 Scenario randomization has become increasingly important for evaluating manipulation generalization, reducing overfitting to fixed laboratory configurations, and creating reproducible training and evaluation splits~\cite{james2020rlbench},~\cite{liu2023libero},~\cite{luo2025fmb},~\cite{li2025easyinsert}. To facilitate reproducible experimentation and large-scale evaluation, we developed a scenario generation tool that automatically creates benchmark configurations. The tool randomizes the position and orientation of hole pieces while respecting the physical constraints imposed by the benchmark layout.

Users can select the desired tolerance level, define the target task type, and specify a random seed that uniquely identifies the generated configuration. For each generated scenario, the tool produces both a visual representation and a machine-readable JSON description containing the complete task specification.

This functionality simplifies experimental replication across laboratories and enables standardized evaluation on shared scenario sets. Moreover, the generated configuration files can be used to create training, validation, and testing splits with controlled levels of novelty, making the benchmark suitable for studying manipulation generalization.

\begin{figure}

\centering
{\label{fig:scenario}
\centering
\includegraphics[width=0.95\linewidth]{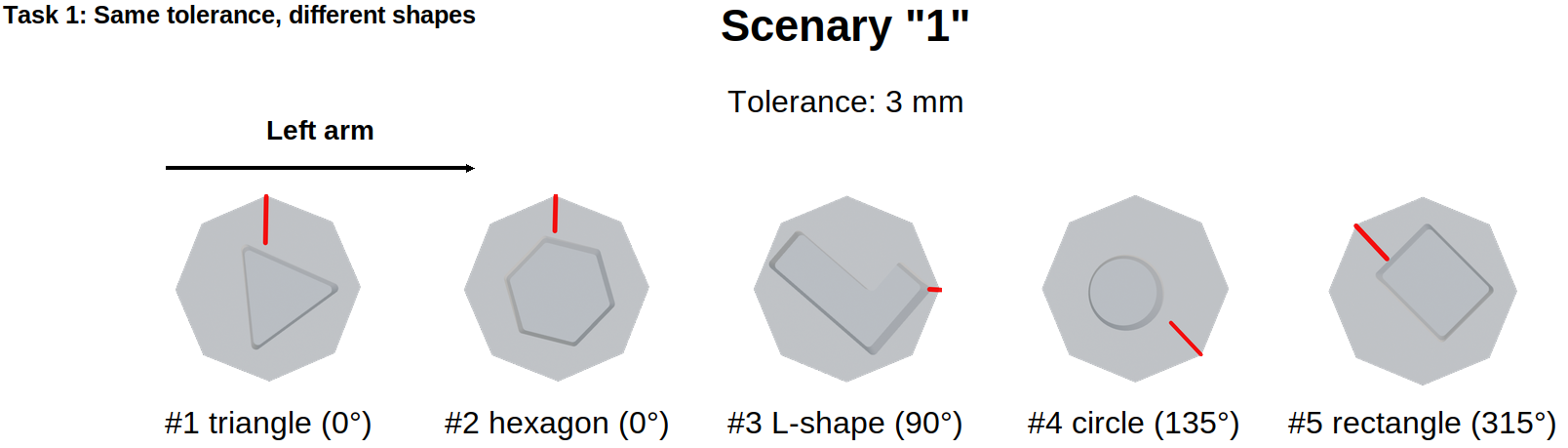}
}
\caption{Scenario for the single-peg insertion task generated by the tool, using 3 mm tolerance holes and random position and orientations.}
\end{figure}

\begin{figure}[t] 
\centering 
\includegraphics[width=0.4\linewidth,angle=90]{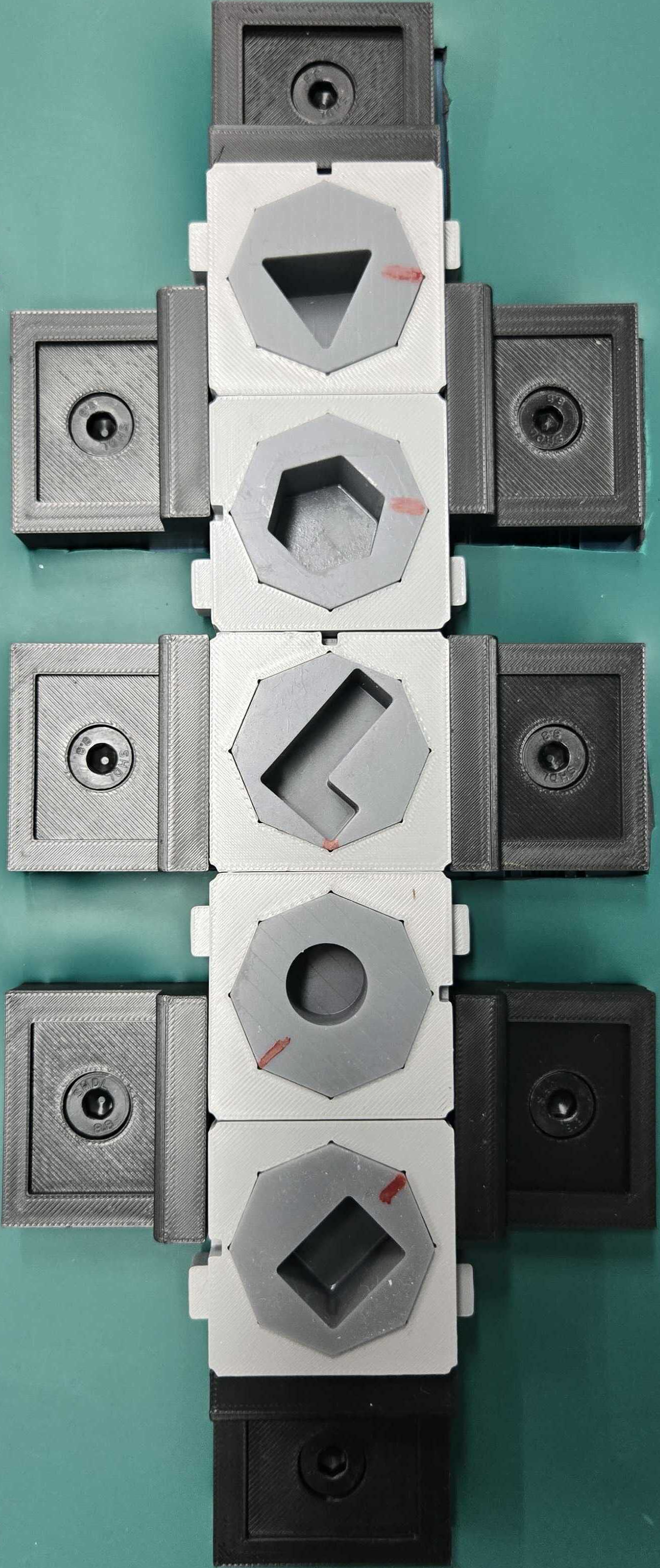} 
\caption{Same scenario replicated using the benchmark pieces in a real-world environment, ready to use for data collection.} 
\label{fig:rwscenario} 
\end{figure}

\section{CONCLUSIONS AND DISCUSSIONS}

The primary contribution of Peg-in-Bench is not the introduction of a new peg-in-hole task, but a framework for generating diverse insertion scenarios from a common set of reusable components. By enabling controlled variation of geometry, tolerance, spatial layout, and assembly structure, the benchmark supports the evaluation of task-level generalization while maintaining reproducible physical conditions. This distinguishes the benchmark from traditional insertion evaluations that focus primarily on performance in fixed task configurations.

The benchmark intentionally focuses on task-level variation while keeping factors such as material properties, friction, illumination, and sensing conditions approximately constant. This design choice allows the influence of task configuration on insertion performance to be studied in isolation. However, it also limits the benchmark's ability to evaluate robustness against perception degradation, manufacturing variability, or changing contact properties. Additionally, the current fixation system was designed around a Vention aluminum profile table, which may require alternative mounting solutions for adoption in other laboratory environments.

We presented a reconfigurable peg-in-hole benchmark for evaluating task generalization in high-precision insertion. The benchmark enables the systematic creation of diverse insertion and assembly scenarios from a common set of physical components and is accompanied by a scenario generation tool that supports reproducible experimentation and standardized evaluation.

\subsection{Future work}

Future work will focus on establishing baseline results, defining standardized evaluation protocols, extending support for more complex assembly and multi-arm manipulation tasks, and improving benchmark accessibility through additional mounting solutions and component refinements. We believe Peg-in-Bench can serve as a common testbed for studying robustness, adaptability, and generalizable robotic insertion and assembly skills across a broad range of manipulation approaches.





\section*{}

\section*{ACKNOWLEDGMENT}

This paper is based on results obtained from a project, JPNP25015, commissioned by the New Energy and Industrial Technology Development Organization (NEDO). 

\balance

\bibliographystyle{IEEEtran} 
\bibliography{IEEEabrv, references}

\end{document}